\documentclass[]{llncs}
\usepackage{pifont}
\usepackage{color}
\usepackage{bbm}
\usepackage{verbatim}
\usepackage{makeidx}  
\usepackage{graphicx}
\usepackage{amssymb,bm}
\usepackage{amsmath}
\usepackage{booktabs}
\usepackage[misc]{ifsym}
\usepackage{multirow}
\usepackage{bbding}
\usepackage{array}
\usepackage{float}
\usepackage{algorithm}
\usepackage{algorithmic}
\usepackage{enumerate}
\newcommand{\etal}{\textit{et al}. }
\newcommand{\ie}{\textit{i}.\textit{e}., }
\newcolumntype{C}[1]{>{\centering\let\newline\\\arraybackslash}m{#1}}

\begin{document}
	\frontmatter          
	%
	%
	\mainmatter              
	\title{A Macro-Micro Weakly-supervised Framework for AS-OCT Tissue Segmentation}
	
	\author{Munan Ning\inst{1,2} \and Cheng Bian\inst{1} \Letter \and
    Donghuan Lu\inst{1} \and Hong-Yu Zhou\inst{1} \and Shuang Yu\inst{1} \and Chenglang Yuan\inst{1} \and Yang Guo\inst{2} \and Yaohua Wang\inst{2} \and Kai Ma\inst{1} \and Yefeng Zheng\inst{1}}
    \institute{
    Tencent Jarvis Lab, Tencent Shenzhen, China \\
    \and
    National University of Defense Technology, China\\
	\email{tronbian@tencent.com} \\
    }
    
	
	\maketitle 
	\date{}
	\hyphenpenalty=5000
	\tolerance=1000
	
	\begin{abstract}
	Primary angle closure glaucoma (PACG) is the leading cause of irreversible blindness among Asian people. Early detection of PACG is essential, so as to provide timely treatment and minimize the vision loss. In the clinical practice, PACG is diagnosed by analyzing the angle between the cornea and iris with anterior segment optical coherence tomography (AS-OCT).
	The rapid development of deep learning technologies provides the feasibility of building a computer-aided system for the fast and accurate segmentation of cornea and iris tissues. However, the application of deep learning methods in the medical imaging field is still restricted by the lack of enough fully-annotated samples. In this paper, we propose a novel framework to segment the target tissues accurately for the AS-OCT images, by using the combination of weakly-annotated images (majority) and fully-annotated images (minority).
	The proposed framework consists of two models which provide reliable guidance for each other. In addition, uncertainty guided strategies are adopted to increase the accuracy and stability of the guidance. Detailed experiments on the publicly available AGE dataset demonstrate that the proposed framework outperforms the state-of-the-art semi-/weakly-supervised methods and has a comparable performance as the fully-supervised method. Therefore, the proposed method is demonstrated to be effective in exploiting information contained in the weakly-annotated images and has the capability to substantively relieve the annotation workload.
	

	

    
	\end{abstract}	
	
	\section{Introduction}
    Glaucoma is the leading cause of irreversible vision loss world-widely that is predicted to affect more than 100 million people by year 2040 \cite{tham2014global}. 
    Primary angle closure glaucoma (PACG), as a major subtype of glaucoma, develops when the angle between the iris and cornea is closed or narrowed, resulting in the blockage of drainage canals and sudden rise in intraocular pressure \cite{sun2017primary}. 
    In the clinical practice, the anterior segment optical coherence technology (AS-OCT)~\cite{radhakrishnan2001real} is widely utilized to obtain both quantitative and qualitative information on the anatomical structures of cornea and iris for the PACG diagnosis~\cite{fu2017segmentation,li2012anterior,niwas2016automated,nolan2007detection}.
    However, manual analysis of each image is laborious and requires professional knowledge. Although the rapid development of deep learning technologies reveals the feasibility of fully automatic anatomical structure segmentation with high accuracy~\cite{fu2018multi}, it still requires a large quantity of images with pixel-wise annotations for the related structures, which is time-consuming and expertise-demanding.

    
    To alleviate the intensive annotation workload of clinicians, a lot of efforts have been made on semi-/weakly-supervised segmentation~\cite{cui2019semi,hung2018adversarial,kervadec2019constrained,perone2018deep,tang2018normalized,tang2018regularized,yu2019uncertainty}. The semi-supervision based methods aim to extract information from a large amount of unlabeled images with the assistance of some fully-annotated images or samples. For example, Perone \etal~\cite{perone2018deep} proposed a semi-supervised teacher-student framework, which leveraged the supervised knowledge learned from the teacher model to improve the segmentation performance of the student model. Yu \etal~\cite{yu2019uncertainty} further adopted the uncertainty information to the teacher-student model to fully exploit the information of the unlabeled data by following the prediction consistencies under different perturbations. Hung \etal~\cite{hung2018adversarial} proposed an adversarial based strategy, which introduced a new discriminator to predict the confidence map for utilizing the information of unlabeled images. However, current semi-supervised methods still require a considerable quantity of fully-annotated images for a satisfactory performance. Another strategy is to improve the workload efficiency by adopting weak annotations\footnote[1]{For the rest of paper, full annotation refers to manual label of each pixel, while weak annotation refers to circles, dots, or scribbles denoting the region of interest.} for training. For example, Kervadec \etal~\cite{kervadec2019constrained} introduced a differentiable term into the proposed loss function to impose the soft size constraints extracted from the weak annotations on the target region. Tang \etal~\cite{tang2018normalized,tang2018regularized} proposed to attain better performance by jointly optimizing the normalized cut with a deep learning model and CRFs for the weakly-supervised task. Although these weakly-supervised methods might relieve the annotation workload to some extent, their segmentation could be error-prone due to the lack of sufficient pixel-wise annotation information. In the clinical practice, apart from a large number of weakly-annotated samples, there is also a small number of full annotations, which might be combined together and employed to improve the model's performance.
    
    To address the above issues of semi-/weakly-supervised learning, an intuitive solution is to integrate both the fully-annotated images and the weakly-labeled samples into the training process, so that the former images can provide accurate pixel-wise tutorial while the latter ones offer more high-level region proposals for segmentation. 

    In this paper, we propose an uncertainty-aware macro-micro (UAMM) framework for the segmentation of the cornea and iris with a few fully-annotated data and a relatively large number of weakly-labeled samples. The network of the proposed UAMM approach consists of two main components with two flows: the macro model with the microscopic flow and the micro model with the macroscopic flow. Unlike the teacher-student framework in which only the teacher model provides guidance to the student model, the macro model and the micro model in the proposed framework offer information for each other to achieve better segmentation performance. Specifically, the macro model utilizes the weakly-labeled samples to learn segmentation proposals to induce the semantic clues for the optimization of the micro model (a.k.a, microscopic flow), while the micro model employs fully-annotated images to present pixel-wise tutorial to guide the learning process of the macro model (a.k.a, macroscopic flow).
    The main contributions of this study are four folds:
    \begin{itemize}
    \item[1)] We propose a novel weakly-supervised methodology for the segmentation of cornea and iris in the AS-OCT images, which outperforms state-of-the-art semi-/weakly-supervised methods and achieves comparable performance as the fully-supervised network.
    \item[2)] Besides the informative features distilled from the weakly-labeled samples, we propose to add the macroscopic flow from the micro model to provide pixel-wise guidance for the optimization of the macro model.
    \item[3)] Other than pixel-wise annotation information learned from the fully-annotated images, the microscopic flow from the macro model is designed to offer more high-level semantic information for the training of the micro model.
    \item[4)] We propose to introduce uncertainty guidance strategies into the microscopic flow and macroscopic flow for more accurate and stable guidance.
    \end{itemize}

	\section{Method}
\begin{figure}[ht]
    \centering
    \includegraphics[width=1.0\columnwidth]{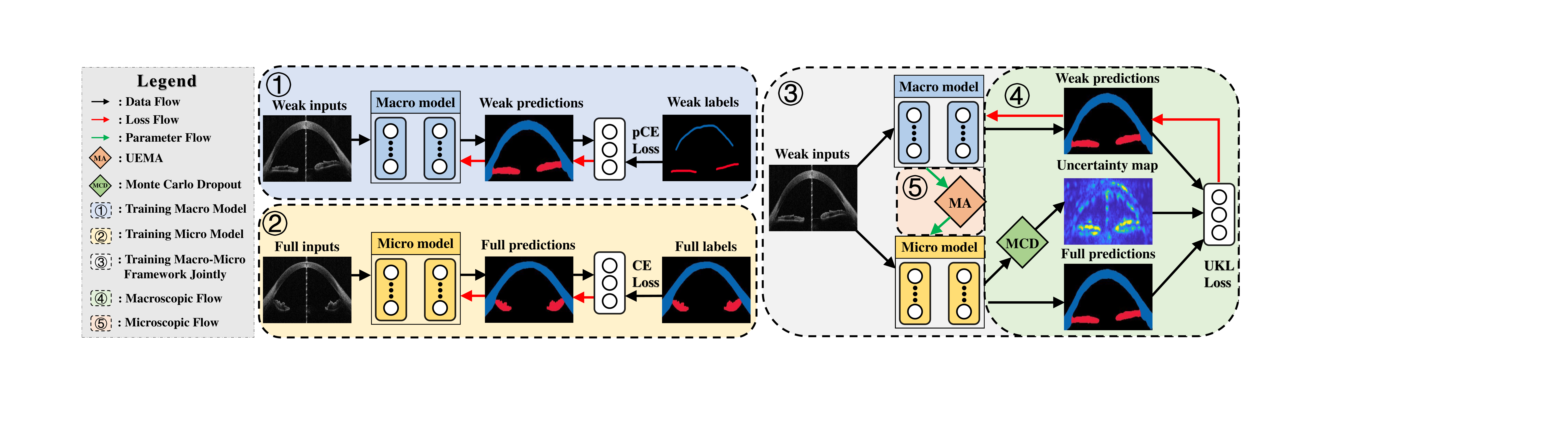}
    \caption{The framework of our uncertainty-aware micro-macro framework. We only use full annotations in stage \ding{173}, while weakly-labeled images in the other.} 
    \label{fig:framework}
    \end{figure}

    Fig.~\ref{fig:framework} displays the diagram of the proposed UAMM framework, which consists of the micro model and the macro model. Both models have the same network architecture, \ie DeepLabV3+~\cite{chen2018encoder}, with different parameters. The proposed framework is optimized via a two-stage training strategy. 
    In the first stage, the two models are trained individually using the fully-annotated images and weakly-labeled samples, i.e., the individual training stage, marked as \ding{172} and \ding{173} in Fig.~\ref{fig:framework}. 
    In the second stage, the two models are trained jointly using only the weakly-labeled samples, \ie the joint training stage marked as \ding{174} in Fig.~\ref{fig:framework}, which provide guidance (the macroscopic and microscopic flows, marked as \ding{175} and \ding{176}) for each other to achieve better segmentation performance. 
    To prevent potential misleading of the incorrect information, uncertainty guidance strategies are proposed to provide more accurate and stable guidance for the model training procedure. To clarify notations, 
    $x \in \mathbb{R}^{H \times W \times 3}$ denotes the input image, where $H,W$ and $3$ represent the height, width and three channels of the input RGB image, respectively; $y^{s}, y^{w}\in\{0,1\}^{H \times W \times C}$ stand for the $C$-way full and weak annotations, respectively; $f$, $\theta$ and $m$ indicate the non-learning transformation of DeepLabV3+, the model parameters and model output, respectively.
    
    \subsection{Loss Functions for the Macro and Micro Model}
    In the first stage, the macro model and micro model are trained separately, \ie the macro model is optimized with the weakly-labeled samples, while the micro model is trained with the fully-annotated images. Specifically, suppose there are $N$ fully-annotated images denoted as $\mathcal{D}_{s}=\left\{\left(x_{i}, y^{s}_{i}\right)\right\}_{i=1}^{N}$, and $M$ weakly-labeled samples represented by $ \mathcal{D}_{w}=\left\{\left(x_{j}, y^{w}_{j}\right)\right\}_{j=1}^{M}$. The loss function for each model in the individual training stage is defined as: 
        \begin{equation}
        \label{eq:loss_for_micro}
            \mathcal{L}_{micro} (x_i)=-\frac{1}{K \times C} \sum_{k=1}^{K} \sum_{c=1}^{C} y_{i}^{s}(k,c) \log m^s_i(k,c)
        \end{equation}
        \begin{equation}
        \label{eq:loss_for_macro}
        \mathcal{L}_{macro} (x_j)=-\frac{1}{\sum_{k=1}^{K} s_{j}(k) \times C} \sum_{k=1}^{K} \sum_{c=1}^{C} s_{j}(k) \cdot y_{j}^{w}(k,c) \log m^w_j(k,c)
        \end{equation}
    where $s_j \in\{0,1\}^{H \times W \times 1}$ is the binary indicator denoting the weakly-annotated pixels; $k$ iterates over all locations with $K$ = $H\times W $ and $c$ iterates over $C$ classes; $m^s_i=f(x_{i}; \theta^{s})$ and $m^w_j=f(x_{j};\theta^{w})$ represent the outputs of the micro model and macro model, respectively.
    
    Eq.~\ref{eq:loss_for_micro} represents the vanilla cross-entropy loss~\cite{tang2018normalized} for the micro model, while Eq.~\ref{eq:loss_for_macro} denotes the partial-cross-entropy (pCE) loss~\cite{tang2018normalized} for the macro-model. The pCE loss only considers the weak label proposals and the relevant regions during the training process, and thus can discourage the probability of mistakenly classifying the unlabeled pixels as the background.
    
    \subsection{Uncertainty-aware KL Loss for the Macroscopic Flow}
    Because pixel-wise labels are not available for the weakly-labeled images, the macro model trained on them can hardly deliver satisfactory segmentation performance. In the second stage, to further improve the accuracy, we utilize the output of the micro model to guide the optimization of the macro model. 
    
    Specifically, we adopt the KL-divergence loss between the output of the two models to fine-tune the macro model. Despite the capability of KL-divergence to align the distributions of two models, the potential mistake of the micro model can result in inaccurate tutorials and mislead the optimization of the macro model. Therefore, we propose to use the uncertainty map to select the reliable pixels for guidance. By using the Monte Carlo dropout (MCD) method~\cite{gal2016dropout}, the uncertainty map can be easily inferred, which serves as an indicator of the reliability of the model's prediction. Specifically, we modify the micro network with several dropout layers, and then repetitively perform the forward pass $T$ times to obtain $T$ Monte Carlo samples $\left\{p_{t}\right\} ^{T}_{t=1}$, where $p_{t}^{c} \in \mathbb{R}^{H \times W \times C}$ denotes the softmax probability map of the $c^{th}$ class at the $t^{th}$ forward pass. Because the variance of Monte Carlo samples can be treated as an approximation of the epistemic uncertainty~\cite{smith2018understanding}, the uncertainty map $U$ of the micro model can be formulated as:
    \begin{equation}
        \label{eq:calculate_U}
    	\mu_c=\frac{1}{T}\sum_{t=1}^{T}{p_{t}^{c}}\quad\quad {\rm and} \quad\quad U=\frac{1}{T \times C}\sum_{t=1}^{T}\sum_{c=1}^{C}(p_{t}^c-\mu_c)^2.
    \end{equation}
    
   Furthermore, an empirical threshold $\tau$ is applied on the uncertainty map to obtain a binary indicator map, in which the positive values represent the reliable pixels. Then, the element-wise multiplication is performed between the KL-loss and the binary indicator map to select the reliable loss for back-propagation. 
   Therefore, for the microscopic flow in the joint training stage, the macro model can be updated via the uncertainty guided KL loss, as defined below:
        \begin{equation}
        \begin{aligned}
            \mathcal{L}_{UKL}&=\frac{\mathbb{I}(U<\tau) \cdot \mathcal{L}_{KL}\left(m^w_j||m^s_i\right)}{\sum\mathbb{I}(U<\tau)}\\
            &=\frac{1}{\sum_{k=1}^{K} \mathbb{I}(U(k)<\tau) \times C}
            \sum_{k=1}^{K} \sum_{c=1}^{C} \mathbb{I}(U(k)<\tau) \cdot m^w_j(k,c) \log \left(\frac{m^w_j(k,c)}{m^s_i(k,c)}\right).
        \end{aligned}
        \label{eq:microscopic_flow}
        \end{equation}
    where $U$ denotes the uncertainty map, $\mathbb{I}(\cdot)$ represents the binary map and the threshold $\tau$ is set to $0.5$ for all the experiments. Note that only weakly-labeled images are used in this step, because the micro model has extremely high confidence for the fully-annotated images, which has already been used to train the model in the first stage.

    \subsection{Uncertainty-aware EMA as the Microscopic Flow}
    As previously stated, the micro model is first trained with the fully-annotated images. Despite the fact that the fully-annotated images contain informative pixel-wise annotation, optimization with a limited number of samples can easily result in overfitting and deteriorate the generalization capability of the model. 
    Therefore, in the second stage, we use the segmentation proposals learned from the macro model to induce the semantic clues for the micro model.
    
    Unlike in the macroscopic flow where the output of the micro model can be directly used as the tutorial, the output of the macro model trained with weakly-labeled samples may not be accurate enough to be used for guidance. Yu \etal~\cite{yu2019uncertainty} proposed an asynchronous updating solution for two collaborative models, \ie the exponential moving average (EMA) mechanism, based on the idea that the weights of the model would contain implicit information of the inference evidence. In this work, the weights of the macro model contain critical information learnt from the weakly-labeled regions and could be useful for the training of the micro model. However, adopting the classic EMA strategy to partially update the micro model with the weights of the macro model requires a predefined updating rate, which may not be the optimal solution. Instead, we propose an uncertainty-aware exponential moving average (UEMA) mechanism for the microscopic flow. $\theta^{s}$ and $\theta^{w}$ are used to represent the weight parameters of the micro and macro model, respectively. The proposed UEMA in the joint training stage can be summarized as:
        \begin{equation}
         \theta^{s}=  \alpha\theta^{s}+(1-\alpha)\theta^{w} 
         \quad\quad {\rm and} \quad\quad 
         \alpha=\frac{\sum_{k=1}^{K} \mathbb{I}\left(U(k)<\tau\right)}{\sum_{k=1}^{K} \mathbbm{1}},
        \end{equation}
    where the $\mathbbm{1} \in\{1\}^{H \times W}$ denotes the unit map with the same shape as $U$. Note that $U$ represents the uncertainty map the same as in Eq.~\ref{eq:microscopic_flow}. The updating rate $\alpha$ is calculated by dividing the sum of uncertainty binary map $\mathbb{I}(U(k)<\tau)$ with the sum of $\mathbbm{1}$. It is used to control the updating rate of UEMA. The less certain the micro model is, the more its parameters are going to be affected by the macro model. Through this asynchronous updating strategy, the segmentation proposal learnt by the macro model can effectively guide the micro model towards better generalization ability with adaptive updating rates.

	\section{Experiment}
	\paragraph{\textbf{Experimental setup}}
	The proposed method is evaluated on a publicly available dataset: the Angle closure Glaucoma Evaluation (AGE) Challenge~\cite{petb-fy10-19}, which provides 3200 AS-OCT images with the dimension of $998\times 2130$ pixels. The original challenge dataset provides annotation for the angle closure classification label and location of the scleral spur. In order to further realize the quantitative analysis of iris and cornea, we have the two key tissues manually re-annotated by experienced  ophthalmologists, and offered two types of annotations, \ie the full annotation and the weak annotation.
	Pixel-wise masks of iris and cornea are provided by the full annotation, meanwhile, for the weak annotation, line strokes inside the tissues are marked. It is worth mentioning that the original PACG classification problem is reformulated to the tissue segmentation problem, therefore we do not use the original annotation in this work.
	
	We randomly select $60\%$ of the images for training, $20\%$ for evaluation and $20\%$ for test (only full annotations are used for evaluation and test). All the images and the corresponding annotations are resized to 240 $\times$ 512 pixels, and the image intensities are normalized into the range of [-1, 1]. The framework is implemented with PyTorch on an NVIDIA Tesla P40 GPU. We utilize the SGD optimizer with $weight\ decay = 0.0005$ and $momentum = 0.9$ to update the network parameters. The batch size is set to 4 for both micro and macro models. Dice coefficient (Dice, represented with percentage) and average distance of boundaries (ADB, represented with millimeter)~\cite{bian2018pyramid} are used as the evaluation criteria. Higher Dice and lower ADB imply better segmentation performance. For convenience, we denote Dice1/ADB1 and Dice2/ADB2 as the evaluation metrics of the cornea and the peripheral iris in this work. 
	
	

    \begin{table*}[ht]
    \caption{Ablation studies on the proposed modules and annotation partition.
         \label{table:ablation}}
    	\centering
    	\renewcommand{\arraystretch}{0.95}
    	\scalebox{0.60}{
    	\large
    	\begin{tabular}{C{3cm}|C{1.5cm}|C{1.5cm}|C{1.5cm}|C{1.5cm}|C{3.0cm}|C{3.0cm}|C{1.5cm}|C{1.5cm}}	
	
    		\toprule[2pt]
            \multirow{3}{*}{\bf{Methods}}&\multicolumn{4}{C{6cm}|}{\multirow{1}{*}{\shortstack{\bf{Combination}}}}
     		&\multicolumn{2}{C{6cm}|}{\bf{Annotation Composition}}& \multicolumn{2}{C{3cm}}{\bf{Ave Metric}}\\
     		\cline{2-9}	
     		&\multicolumn{1}{C{1.5cm}|}{\multirow{2}{*}{\shortstack{\bf{Micro}}}}
     		&\multicolumn{1}{C{1.5cm}|}{\multirow{2}{*}{\shortstack{\bf{Macro}}}}
     		&\multicolumn{1}{C{1.5cm}|}{\multirow{2}{*}{\shortstack{\bf{Macro}\\\bf{Flow}}}}
     		&\multicolumn{1}{C{1.5cm}|}{\multirow{2}{*}{\shortstack{\bf{Micro}\\\bf{Flow}}}}
    		&\multicolumn{1}{C{3cm}|}{\multirow{2}{*}{\shortstack{\bf{Full}\\\bf{Annotation}}}} &\multicolumn{1}{C{3cm}|}{\multirow{2}{*}{\shortstack{\bf{Weak}\\\bf{Annotation}}}} 	&\multicolumn{1}{C{1.5cm}|}{\multirow{2}{*}{\shortstack{\bf{Dice}}}}
     		&\multicolumn{1}{C{1.5cm}}{\multirow{2}{*}{\shortstack{\bf{ADB}}}}\\
     	    &\quad&\quad&\quad&\quad&\quad&\quad&\quad&\quad\\
            \cline{1-9}
            \multirow{3}{*}{\shortstack{\\\\\\\\\\\\\\\\\bf{Module}\\\\\\\bf{ablations}}}
            &\checkmark&\quad&\quad&\quad&1\%&99\%&82.13&2.85\\
            &\checkmark&\checkmark&\quad&\quad&1\%&99\%&83.55&1.87\\
            &\checkmark&\checkmark&$^{*}$&\quad&1\%&99\%&86.70&1.15\\
            &\checkmark&\checkmark&\checkmark&\quad&1\%&99\%&89.35&0.66\\
            &\checkmark&\checkmark&\checkmark&$^{*}$&1\%&99\%&90.12&0.43\\
            &\checkmark&\checkmark&\checkmark&\checkmark&1\%&99\%&\textbf{91.64}&\textbf{0.30} \\
            \cline{1-9}
            \multirow{4}{*}{\shortstack{\\\bf{Annotation}\\\\\\\bf{compositions}}}
            &\checkmark&\checkmark&\checkmark&\checkmark&5\%&95\%&92.48&0.27\\
            &\checkmark&\checkmark&\checkmark&\checkmark&10\%&90\%&92.60&0.26\\
            &\checkmark&\checkmark&\checkmark&\checkmark&25\%&75\%&93.06&0.22\\
            &\checkmark&\checkmark&\checkmark&\checkmark&50\%&50\%&\textbf{93.42}&\textbf{0.20}\\

    		\bottomrule[2pt]
    		\multicolumn{9}{l}{\footnotesize{$^{*}$: The same flow being adopted in this study without uncertainty assistance.}} \\
    		\multicolumn{9}{l}{\footnotesize{Macro Flow: adding macroscopic flow with our uncertainty-aware KL loss.}} \\
    		\multicolumn{9}{l}{\footnotesize{Micro Flow: adding microscopic flow with our uncertainty-aware EMA mechanism.}} \\
    	\end{tabular}}
    \end{table*}
    
    \paragraph{\textbf{Ablation study. }}
    To demonstrate the effectiveness of the proposed modules, we conduct ablation studies as well as experiments with different annotation composition. As shown in Table~\ref{table:ablation}, the performance has improved around 5.80\% and 2.29\% in average Dice by adding the macroscopic and microscopic flow, respectively. In order to evaluate the effect of the proposed uncertainty strategies, the results of flows without uncertainty are presented as well, \ie marked by the asterisk symbol. To be more specific, we use the conventional EMA for the macroscopic flow and the KL-loss for the microscopic flow directly. As expected, the result without uncertainty shows inferior performance (2.65\% lower for macroscopic flow and 1.52\% lower for microscopic flow, respectively), demonstrating that the proposed uncertainty strategies can improve the effectiveness of the tutorials. 
    To evaluate the stability of the proposed method, we conducted additional experiments with different percentages of fully-annotated images. As expected, the more fully-annotated images we utilize, the better performance the method achieves, indicating that the proposed method can exploit the information from full annotations as well. 
    
    \begin{figure}[b]
	\centering
	\includegraphics[width=0.75\columnwidth]{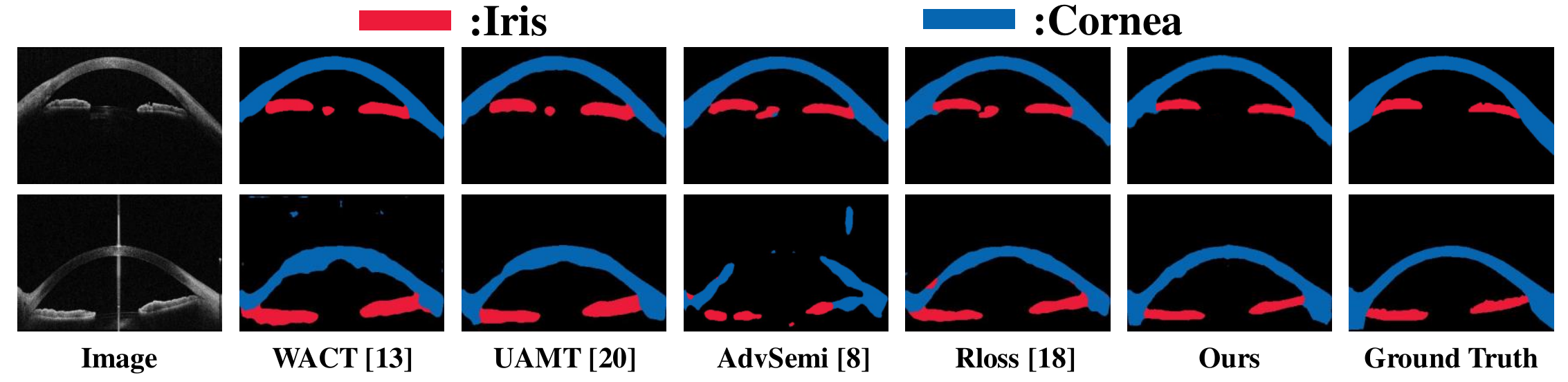}
	\caption{Visualization of the segmentation results by different methods and ours.}
	\label{fig:visualization}
    \end{figure}
    
    \paragraph{\textbf{Comparison with State-of-the-art }}
    As illustrated in Table~\ref{table:sota}, the two columns within the annotation composition represent the percentages of fully-annotated and weakly-labeled images used for training. The results of state-of-the-art semi-/weakly-supervised methods, including WACT~\cite{perone2018deep}, UAMT~\cite{yu2019uncertainty}, AdvSemi~\cite{hung2018adversarial}, and CRF-rloss~\cite{tang2018regularized}, are presented for comparison. In the training set for the proposed UAMM method, only 1\% images are fully-annotated while the rest 99\% samples are weakly-labeled. 
    For the semi-supervised methods, \ie WACT, UAMT and AdvSemi, generally weakly-annotated samples will not be utilized in their studies. Similarly, the full-annotated samples are not used in the weakly-supervised studies, \ie CRF-rloss, either. For a fair comparison, both of full and weakly-annotated samples will be integrated in the training procedure and provide two versions of results, so as to keep the model comparison under the same evaluation criteria. 
     Oracle indicates using only the micro model, \ie a single DeepLabV3+ network~\cite{chen2018encoder}. As the baseline method, Oracle has been applied on four training sets with different percentages of fully-annotated images and weakly-labeled samples, as denoted in Row 2. With the same training data setup, the proposed UAMM method has achieved the best performance among these methods, with 91.64\% in average Dice score and 0.3 in ADB. Furthermore, the evaluation metrics of UAMM are close to the metrics of fully-annotated trained Oracle (only 2.01\% lower on average Dice), demonstrating that the proposed method can exploit segmentation guidance from the weak annotations. The visualization of representative examples is displayed in Fig.~\ref{fig:visualization}.
    
    \begin{table*}[t]
    \caption{
        Quantitative comparison with the state-of-the-art semi-/weakly-supervised learning algorithms.
        \label{table:sota}}
    	\centering
    	\renewcommand{\arraystretch}{0.95}
    	\scalebox{0.60}{
    	\large
    	\begin{tabular}{c|C{3.0cm}|C{3.0cm}|C{1.5cm}|C{1.5cm}|C{1.5cm}|C{1.5cm}|C{1.5cm}|C{1.5cm}}	
	
    		\toprule[2pt]
     		\multirow{3}{*}{\bf{Method}} &
     		\multicolumn{2}{C{6cm}|}{\bf{Annotation Composition}}& \multicolumn{6}{C{9.5cm}}{\bf{Metric}}\\
     		\cline{2-9}		
    		&\multicolumn{1}{C{3cm}|}{\multirow{2}{*}{\shortstack{\bf{Full}\\\bf{Annotation}}}} &\multicolumn{1}{C{3cm}|}{\multirow{2}{*}{\shortstack{\bf{Weak}\\\bf{Annotation}}}} 	&\multicolumn{3}{C{4.8cm}|}{\bf{Dice}}
     		&\multicolumn{3}{C{4.8cm}}{\bf{ADB}}\\
     		\cline{4-9}
    		&&&\textbf{Dice1}&\textbf{Dice2}&\textbf{Ave}&\textbf{ADB1}&\textbf{ADB2}&\textbf{Ave}\\
    		\cline{1-9}
    		\multirow{4}{*}{\shortstack{\bf{Oracle}}}
    		&100\%	&0\%	&\textbf{95.71}	&\textbf{91.59}	&\textbf{93.65}	&\textbf{0.13}	&\textbf{0.21}	&\textbf{0.17}
\\
    		&0\%	&100\%	&55.14	&35.03	&45.09	&9.30	&13.62	&11.46
\\
    		&1\%	&0\%	&78.83	&68.64	&73.73	&5.79 	&6.29 	&6.04
\\
    		&1\%	&99\%	&83.71	&80.55	&82.13	&2.58 	&3.11 	&2.85 
\\
    		\cline{1-9}
    		\multirow{2}{*}{\bf{WACT}~\cite{perone2018deep}}
    		&1\%	&0\%	&51.63	&25.89	&38.76	&9.74 	&19.97 	&14.86 \\
    		&1\%	&99\%	&84.74	&83.13	&83.94	&1.19	&0.77	&0.98\\
    		\cline{1-9}
    		\multirow{2}{*}{\bf{UAMT}~\cite{yu2019uncertainty}}
    		&1\%	&0\%	&86.59	&64.23	&75.41	&2.53 	&5.79 	&4.16 \\
    		&1\%	&99\%	&88.58	&85.06	&86.82	&0.47	&0.73	&0.60\\
    		\cline{1-9}
    		\multirow{2}{*}{\bf{AdvSemi}~\cite{hung2018adversarial}}
    		&1\%	&0\%	&84.02	&69.49	&76.75	&2.88 	&6.11 	&4.50 \\
    		&1\%	&99\%	&88.36	&83.33	&85.85	&3.19	&2.02	&2.6\\
    		\cline{1-9}
    		\multirow{2}{*}{\bf{CRF-rloss}~\cite{tang2018regularized}}
    		&0\%	&100\%	&86.37	&83.97	&85.17	&1.09 	&0.87 	&0.98 \\
    		&1\%	&99\%	&93.44	&83.26	&88.35	&0.32	&0.92	&0.62 \\
    		\cline{1-9}
            \multirow{1}{*}{\bf{UAMM}}
    		&1\%	&99\%	&\textbf{93.68}	&\textbf{89.60}	&\textbf{91.64}	&\textbf{0.27} 	&\textbf{0.32} 	&\textbf{0.30}\\
    		\bottomrule[2pt]
    	\end{tabular}}
    \end{table*}

	\section{Conclusion}
	In this work, we proposed a macro-micro weakly-supervised framework to tackle the problem of cornea and iris segmentation for the AS-OCT images. Specifically, an uncertainty-aware KL loss is designed for the macroscopic flow to assist the training of the macro model by the prediction priors from the micro model. Then, the microscopic flow is obtained with an uncertainty-aware moving average mechanism, which updates the micro-model by gradually involving the weights of the macro model. Our approach outperformed state-of-the-art semi-/weakly-supervised methods on the cornea and iris segmentation task for AS-OCT images. In addition, it achieved comparable performance by using only 1\% of fully-annotated data with that of DeepLabV3+ using all fully-annotated images.
	
	\section{Acknowledgment}
	This work was funded by the Key Area Research and Development Program of Guangdong Province, China (No. 2018B010111001), National Key Research and Development Project (No. 2018YFC2000702) and Science and Technology Program of Shenzhen, China (No. ZDSYS201802021814180).
	
	%
	\bibliographystyle{splncs04}
	\bibliography{SemiSeg}
\end{document}